\documentclass[journal]{IEEEtran}
\usepackage{amsthm,amssymb,scrextend,mathrsfs,amsmath,amsfonts}
\usepackage{mathtools}
\usepackage[
    colorlinks=true,
    linkcolor=blue,
    urlcolor=blue,
    citecolor=blue
]{hyperref}
\usepackage{array}
\usepackage{float}
\usepackage[caption=false,font=normalsize,labelfont=sf,textfont=sf]{subfig}
\usepackage{booktabs}
\usepackage{textcomp}
\usepackage{stfloats}
\usepackage{url}
\usepackage{verbatim}
\usepackage{graphicx}
\usepackage{cite}
\usepackage{tabularx}
\usepackage{multirow}
\usepackage[ruled]{algorithm2e}
\usepackage{bm}
\usepackage{threeparttable}

\usepackage{xcolor}

\theoremstyle{definition}

\newcolumntype{C}[1]{>{\centering\arraybackslash}m{#1}}

\begin{document}

%\title{Data-Driven LC Models: A Comprehensive Survey}

\title{A Survey on Data-Driven Modeling of Human Drivers' Lane-Changing Decisions}

\author{Linxuan Huang, Dong-Fan Xie, Li Li, {\it Fellow, IEEE}, Zhengbing He, {\it Senior Member, IEEE}
        % <-this % stops a space
\thanks{This work is supported by the National Natural Science Foundation of China (72288101, 72171018, 72242102). 
(\textit{Corresponding author: Dong-Fan Xie, and Zhengbing He})}% <-this % stops a space
\thanks{L. Huang is with the School of Systems Science, Beijing Jiaotong University, Beijing 100044, China. 
D.-F Xie is with the School of Systems Science, Beijing Jiaotong University, Beijing 100044, China (e-mail: dfxie@bjtu.edu.cn). 
L. Li is with the Department of Automation, BNRist, Tsinghua University, Beijing 100084, China.
Z. He is with Laboratory for Information and Decision Systems, Massachusetts Institute of Technology, Cambridge MA 02139, the United States
(e-mail: he.zb@hotmail.com)}}

\maketitle

\begin{abstract}
Lane-changing (LC) behavior, a critical yet complex driving maneuver, significantly influences driving safety and traffic dynamics. 
Traditional analytical LC decision (LCD) models, while effective in specific environments, often oversimplify behavioral heterogeneity and complex interactions, limiting their capacity to capture real LCD.
%applicability in complex traffic scenarios. 
Data-driven approaches address these gaps by leveraging rich empirical data and machine learning to decode latent decision-making patterns, enabling adaptive LCD modeling in dynamic environments. 
In light of the rapid development of artificial intelligence and the demand for data-driven models oriented towards 
 connected vehicles and 
autonomous vehicles, this paper presents a comprehensive survey of data-driven LCD models, with a particular focus on human drivers’ LC decision-making. It systematically reviews the modeling framework, covering data sources and preprocessing, model inputs and outputs, objectives, structures, and validation methods.
This survey further discusses the opportunities and challenges faced by data-driven LCD models, including driving safety, uncertainty, as well as the integration and improvement of technical frameworks.
%This survey aims to highlight key advancements and existing gaps, with the goal of guiding further research and development in the data-driven LCD modeling.

\end{abstract}


\section{Introduction}

\IEEEPARstart{L}{ane-changing} (LC), as a fundamental driving maneuver, exerts a profound influence on road safety\cite{PANDE_2006_assessment, ZHENG_2010_impact,CHEN_2019_key,LI_2020_shortterm} and traffic flow dynamics\cite{LAVAL_2008_microscopic,ZHENG_2011_freeway,9907780,hou2024precise}. 
Compared to car-following (CF) behavior, LC behavior entails higher collision risks due to its dependency on holistic evaluations of traffic conditions in both the original and target lanes, requiring drivers to navigate multi-criteria decision-making processes.
More specifically, safe LC execution necessitates gaps in the target lane to satisfy collision-avoidance criteria.
Drivers must continuously monitor the real-time states of surrounding vehicles (e.g., velocity, acceleration) and adjust their LC maneuvers in response to unexpected behavioral changes (e.g., sudden deceleration, lane encroachment).
Human drivers' irrational decision-making (e.g., sudden risk-preference shifts) in dynamic environments pose challenges to traditional  LC models based on hypothesis of rational man.
This complex decision-making process significantly impacts traffic safety and traffic efficiency. 
%Consequently, it is crucial to conduct an in-depth exploration of LC behavior and uncover its impacting factors. 
%Subsequently, LC models with outstanding generalization capabilities should be developed.
Consequently, it is essential to deeply explore the mechanisms underlying LC decisions and develop LC models capable of accurately characterizing the decision-making process. 
This effort will provide critical support for trustworthy traffic simulations, dynamic traffic management, and LC decision-making of autonomous vehicles (AVs).

Generally speaking, LC process comprises two components: LC decision (LCD) and LC implementation (LCI) \cite{ZHENG_2014_recent}. 
LCD, as the critical cognitive component that determines the feasibility and safety of a maneuver, has received predominant attention in lane-changing modeling studies.
The modeling of LCD generally focuses on three fundamental research questions: 
(1) What are the key determinants affecting drivers' LCD under dynamic traffic conditions? 
(2) How can we establish a generalized computational framework that accurately captures the nonlinear relationships between influencing factors and LCD behaviors? 
(3) What methodological innovations are required to develop LCD models that demonstrate robust generalization capabilities across diverse driving scenarios?

Analytical LCD models, grounded in mathematical formulations and probabilistic theories, encompass both rule-based reactive models \cite{GIPPS_1986_MODEL,YANG_1996_MICROSCOPIC,AHMED_1996_MODELS, KESTING_2007_GENERAL} and game-theoretic interactive models \cite{KITA_1999_MERGING,WANG_2015_GAME,ALI_2021_CLACD}.
These models, characterized by well-defined physical principles and structured modeling procedures, are typically categorized as analytical approaches.
Once calibrated, these models demonstrate proficiency in executing LC tasks within constrained environments, such as specific roadways. 
However, their focus on abstracting and simplifying driving behavior patterns makes it difficult to explicitly represent latent behavioral characteristics, such as driving style heterogeneity and social collaboration tendencies. 
This limitation significantly restricts their generalizability across diverse traffic scenarios and varying traffic conditions over time.

In contrast, data-driven models leverage real driver behavior data to uncover complex behavioral patterns and latent features through learning-based approaches.
%supervised learning and reinforcement learning (RL) techniques. 
This approach effectively establishes a mapping from diverse inputs to LC behavior, thereby inherently excelling in achieving human-like decision-making capabilities. 
Such models facilitate the transition from group behavioral norms to individualized characteristics, enabling the implementation of dynamic personalized LC strategies and adaptability to complex interactive scenarios in open-road environments. 
Furthermore, advancements in data acquisition technologies and distributed computing frameworks have provided robust foundations for data-driven LCD modeling by enabling the collection and utilization of large-scale, high-precision vehicle trajectory data.

Existing data-driven LC models can be broadly categorized into the following two groups.
Human behavior-oriented models, which are designed to understand and describe human drivers' LC behavior;
AV-oriented models, which aim at enhancing AV performance.
These two groups of models differ fundamentally in objectives, methodologies, and applications.
\begin{itemize}
\item 
Human behavior-oriented models explore the “why” behind lane changes, capturing subjective factors (e.g., aggressive/conservative styles) and stochastic interactions (e.g., unspoken cooperation) \cite{ZHENG_2014_recent}. 
They rely on probabilistic models and supervised learning to predict LC behaviors, ultimately improving the applications such as human-centric warnings in advanced driver assistance systems (ADAS).

\item 
AV-oriented models solve the “how” of lane changes, converting sensor data into optimal actions via various modeling frameworks (e.g., reinforcement learning rewards, game theory). 
They enforce objective-driven decisions (e.g., safety guarantees, traffic efficiency) and integrate with control systems for machine-executable maneuvers \cite{7515222,9579446,Zong31122022,doi103233}.
\end{itemize}

Despite their differences, the two perspectives are interconnected: insights from human behavior models can inform AV algorithms to improve human-like performance, while data from AVs can refine behavior models by offering a broader range of observations and more granular behavioral patterns, especially in complex traffic scenarios. This design emphasizes robustness and adaptability to ensure the safe integration of AVs into traffic systems.

This survey primarily focuses on reviewing data-driven modeling of human drivers' LCD, covering their methodologies, structural frameworks, existing challenges, and emerging trends.
While partially covering LCD models for AVs, the primary aim remains to evaluate their system-level impacts on traffic flow dynamics, such as traffic simulations, rather than delving into vehicle-specific control strategies (e.g., perception systems, real-time motion planning). 
By maintaining this scope, we aim to clarify how to develop data-driven models that capture LCD behaviors, while also acknowledging the increasing interplay between AV technologies and traffic flow modeling.
Given the vastness of the topic, the following is excluded from the scope of this study:
(1) Driver physiological/psychological states (e.g., distracted driving, health abnormalities, and other confounding factors);
(2) Studies that focus solely on vehicle motion without integrating data-driven insights into individual driver behavior;
(3) Moral dilemma scenarios (e.g., trolley dilemmas), which are excluded due to their incompatibility with microscopic driving models.

An increasing number of researchers are acknowledging the benefits of utilizing data-driven methods in understanding and modeling LCD behaviors.
These methods utilize substantial trajectory data to capture complex driving behaviors and enhance models through machine learning (ML) and statistical analysis techniques. 
Despite the extensive literature,  our objective is to shed light on the following issues that have not received sufficient attention and provide a concise technical framework.
\begin{itemize}
\item 
The application of data-driven algorithms in LCD has been extensively reviewed. These algorithms leverage historical and real-time traffic data to model human drivers' LCD effectively. 
The integration of data-driven approaches into LCD modeling represents a potential paradigm shift in the field. 
%However, there is a lack of comprehensive reviews on LCD models from a data-driven perspective, particularly concerning human driver decision-making patterns in dynamic environments.

\item 
This survey provides a comprehensive and thorough review of existing data-driven LCD models. It summarizes the key characteristics of these models, including their data sources, data processing methods, inputs and outputs, structure, objectives, and validation techniques. Each component is described with illustrative examples, and the performance of representative models across these aspects is evaluated.

\item 
Several limitations of current data-driven LCD models are identified, with significant challenges related to robustness, uncertainty, and framework highlighted and discussed. These factors hinder the validation and development of data-driven LCD models while simultaneously creating opportunities for diversified modeling methodologies.

\end{itemize}

The remainder of this survey is organized as follows: Section \ref{Adv-LC-Models} presents the development stages of data-driven LCD models, while Section \ref{Data-driven LCD model} discusses various data-driven LCD models. Section \ref{Principles of DDLC models} summarizes the fundamental principles underlying these models. Section \ref{Opportunities and challenges} addresses the challenges associated with different methods and explores the opportunities in data-driven LCD modeling. Finally, Section \ref{Conclusion} concludes with the main findings.

\section{The evolution of data-driven LCD models}
\label{Adv-LC-Models}

This comprehensive review is primarily conducted through the Web of Science database, employing a systematic search methodology to identify relevant papers. 
The analysis of research developments over the past decade is performed using bibliometric techniques and scientific knowledge mapping approaches. 
The past fifteen years have witnessed unprecedented advancements in data-driven methodologies, particularly marked by the emergence of big data analytics, substantial improvements in algorithms such as convolutional neural networks, and significant enhancements in computational capabilities \cite{KRIZHEVSKY_2017_IMAGENET}. 
Subsequently, data-driven models have progressively been implemented in the domains of LCD modeling studies. 
The present study conducts a statistical analysis of papers and review papers published within the last decade. 
The search strategy employed a predefined set of keywords, as detailed in Table \ref{tab:keywords search}, which yielded 343 relevant publications. 
Furthermore, a cluster analysis was performed on the keywords extracted from these publications to identify prevailing research trends and thematic concentrations.

\begin{table}[!ht]
\caption{\label{tab:keywords search}Set of keywords for literature search.}
\centering
\begin{tabular}{p{6.2cm}p{1.5cm}}
\toprule % 第一道横线
\textbf{Set of search keywords}  & \textbf{Description}  \\
\midrule % 第二道横线 
Topic = ("lane change" OR "lane-changing" OR "LC" OR "change lanes" OR "lane-changing behavior" OR "lane-changing prediction") & LC behavior  \\ 
\vspace{-0.1cm} 
Topic = ("data-driven" OR "data driven" OR "data mining" OR "artificial intelligence" OR "machine learning" OR "deep learning" OR "reinforcement learning" OR "neural networks" OR "predictive modeling" OR "ensemble learning") & Data-driven  \\ 
\bottomrule % 第三道横线
\end{tabular}
\end{table}

\begin{figure*}
\centering
\includegraphics[width=0.8\linewidth]{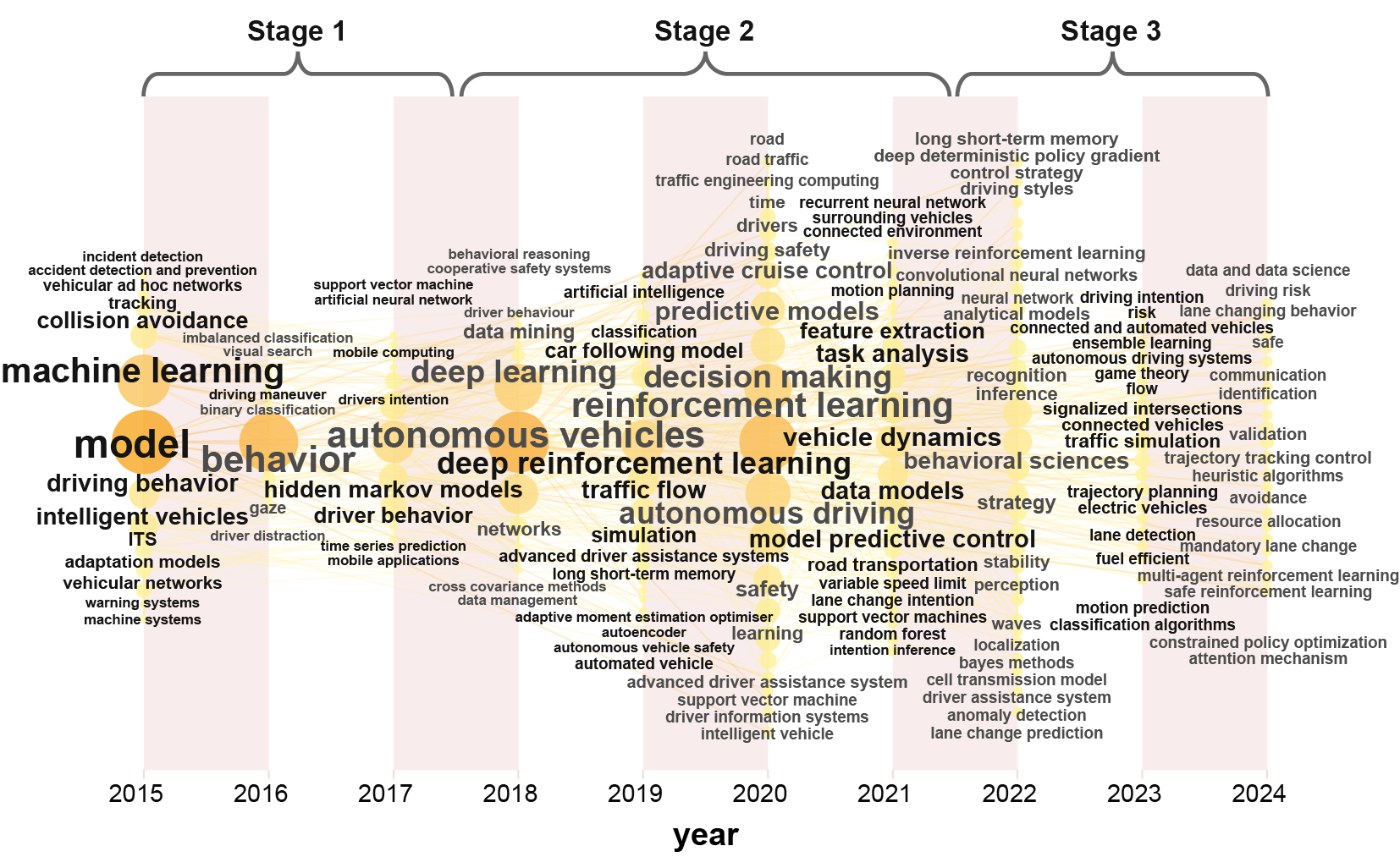}
\caption{\label{fig:Keyword Time Zone}Keyword time zone distribution map of data-driven LC models.}
\end{figure*}

Figure \ref{fig:Keyword Time Zone} illustrates the timeline of the evolution of relationships among sub-themes in the topic of data-driven LCD models, clearly highlighting the research trends. 
Notably, AV technology significantly propels the integration of data-driven approaches in LCD models. 
From the perspective of methodology development, data-driven LCD studies from 2015 to 2024 can be delineated into the following three distinct stages:

\begin{itemize}
\item {\bf Stage 1 (2015-2017):}
This stage marked the transition of LC behavior studies from rule-based and traditional kinematic models to data-driven approaches. 
Researchers began systematically integrating machine learning techniques, such as support vector machines (SVM) and hidden markov models (HMM), to construct models for LCD recognition. 
Concurrently, methodology innovations in intelligent transportation systems (ITS) gradually permeated the domain of LCD. 
Additionally, behavioral science was incorporated into LC research, unveiling the decision-making pattern variations across different driving styles and facilitating the design of personalized LCD strategies.

\item {\bf Stage 2 (2018-2021):}
This stage focused on the methodological advancements of data-driven approaches, with deep learning (DL) and reinforcement learning (RL) emerging as prominent keywords. 
Research during this stage emphasized the construction of more intricate and refined models from multifactorial and multimodular perspectives, demonstrating the superiority of data-driven methods in extracting LC scene features. 
Moreover, the utilization of deep networks established diverse driving styles and environmental interaction paradigms, thereby advancing LC research from a “vehicle-centric” approach to a “human-vehicle-environment coupling” framework.

\item {\bf Stage 3 (2022-2024):}
This stage is characterized by a plethora of diverse keywords, with LCD modeling methods building upon and refining the models developed in Stage 2.
The interaction and integration among the “human-vehicle-environment” triad have become the focal point of LCD models. 
Researchers’ attention has gradually shifted from the precision of behavior prediction to the interactive performance of models, the differentiation of driving styles, scenario adaptability, and the impact on traffic flow. 
LCD research has pivoted towards exploring complex traffic system scenarios, demanding higher levels of model adaptability, generalization capability, and robustness.
\end{itemize}

The evolution of LCD modeling not only reflects technological advancements but also demonstrates a deeper understanding of LCD mechanism and its impacts on traffic flow. 
As technology becomes increasingly sophisticated, the LCD models also expose more issues need to be addressed.

\section{Data-driven LCD models}
\label{Data-driven LCD model}

The core objective of LCD models lies in determining the optimal timing for LC maneuvers to achieve safe and efficient driving.
This research domain not only encompasses the construction of algorithmic frameworks but also requires a comprehensive integration of traffic flow dynamics, road condition information, and the behavioral patterns of surrounding vehicles. 
In the development of data-driven LCD models, supervised learning typically formulates human-like LCDs as classification problems. 
DL methodologies, leveraging their robust feature extraction and learning capabilities, exhibit exceptional performance in tasks related to behavior prediction and intent recognition. 
Furthermore, the essence of LCDs resides in the accurate comprehension of both the actions of neighboring vehicles and the driver’s own operational tasks in highly interactive and dynamic traffic environments. 
Consequently, precise apprehension of the traffic situation constitutes a fundamental prerequisite for effective LCD.

From the perspective of data-driven modeling approaches, namely, classical ML, DL, RL, and data-physics hybrid modeling approach, this section conducts literature review on LCD models.

\subsection{LCD models based on classical ML}

Traditional machine learning methods are renowned for their simplicity and interpretability, making them particularly suitable for scenarios with limited data and well-defined features. 
Among these methods, algorithms such as SVM, gaussian mixture models (GMM), and HMM are widely applied. 
The selection of a specific algorithm typically depends on the characteristics of the problem, the nature of the data, and the desired interpretability and performance.

Specifically, SVM constructs an optimal hyperplane to classify data samples into distinct categories. 
Bayesian networks (BN) are adept at handling uncertainty and managing missing data. 
GMM, as nonparametric probabilistic models, can represent the multimodal distribution characteristics of driving behavior through latent variables and are often embedded as feature preprocessing modules in hybrid architectures, such as driving style classifiers\cite{WANG_2019_LATERAL}. 
For instance, \cite{KUMAR_2013_LEARNING} proposed an LC intention recognition model based on SVM and Bayesian filter, achieving high accuracy in classification of LC and lane keeping (LK) behaviors. 
\cite{LIU_2019_NOVEL} proposed the Gaussian kernel SVM-based LCD model with benefit, safety and tolerance as inputs, significantly enhancing the accuracy of the LC classification. 
\cite{HOU_2014_MODELING} developed a more conservative LCD model by integrating a k-nearest neighbor (KNN) density-based Bayesian classifier with a decision tree model through voting. 
Subsequently, \cite{HOU_2015_SITUATION} further optimized the integration rules by employing ensemble learning methods such as Random Forest and AdaBoost to construct LC classifiers, achieving higher classification accuracy and lower false positive rates, thereby significantly improving the safety of LCDs.

From a theoretical modeling perspective, HMM possess the capability to model time-series data of arbitrary lengths.
By constructing a hidden state transition probability matrix and an observation sequence generation mechanism, HMM can effectively extract latent intention evolution patterns from continuous driving dynamics. \cite{MEYER_2009_PROBABILISTIC} recognized the behaviors of following and overtaking vehicles by applying HMM-GMM. 
\cite{LI_2016_LANE} integrated HMMs inspired by speech recognition models with a Bayesian filter (BF) to discern drivers' LC intentions.
\cite{DAVID_2022_STUDY} developed a state machine approach based on HMM, which effectively improved the performance of accuracy, detection rate, and false alarm rate, while discovering that time-to-collision (TTC) features can enhance recognition performance.

Classical ML algorithms are characterized by their transparent structures and strong interpretability, enabling acceptable prediction accuracy on small to medium-sized datasets. 
However, these shallow architecture models often struggle to simultaneously meet the demands for high prediction accuracy and strong generalization capabilities in complex driving scenarios. 
This is primarily manifested in: 
(1) limited feature extraction capabilities, making it difficult to capture deep semantic information in driving scenarios; 
(2) insufficient model capacity to effectively handle complex relationships in high-dimensional nonlinear feature spaces; 
and (3) constrained generalization performance, leading to performance degradation in new LC scenarios. 
Consequently, classical ML methods are inadequate for high-precision, strongly generalized driving decision-making LC tasks.

\subsection{LCD models based on DL methods}

The advancement of DL algorithms has endowed deep networks with the more powerful capability of mining the correlations and couplings between features.
This improved capability enables deep networks to perform exceptionally well in tasks characterized by strong nonlinearity, implicit rules, and highly coupled features.
It can be stated that DL architectures inherently excel at extracting LC information from human driving datasets, offering natural advantages in achieving human-like LC behavior. 
\cite{XIE_2019_DATADRIVEN} pioneered the development of a LCD model based on deep belief networks (DBN). 
Through comparative experiments with the logit model and backpropagation neural networks, their study demonstrated substantial superiority in LC feature extraction and classification tasks. 
This research not only validates the performance advantages of DL methods in learning LC behaviors but also lays a crucial foundation for subsequent studies in driving behavior modeling.

\begin{figure}
\centering
\includegraphics[width=1\linewidth]{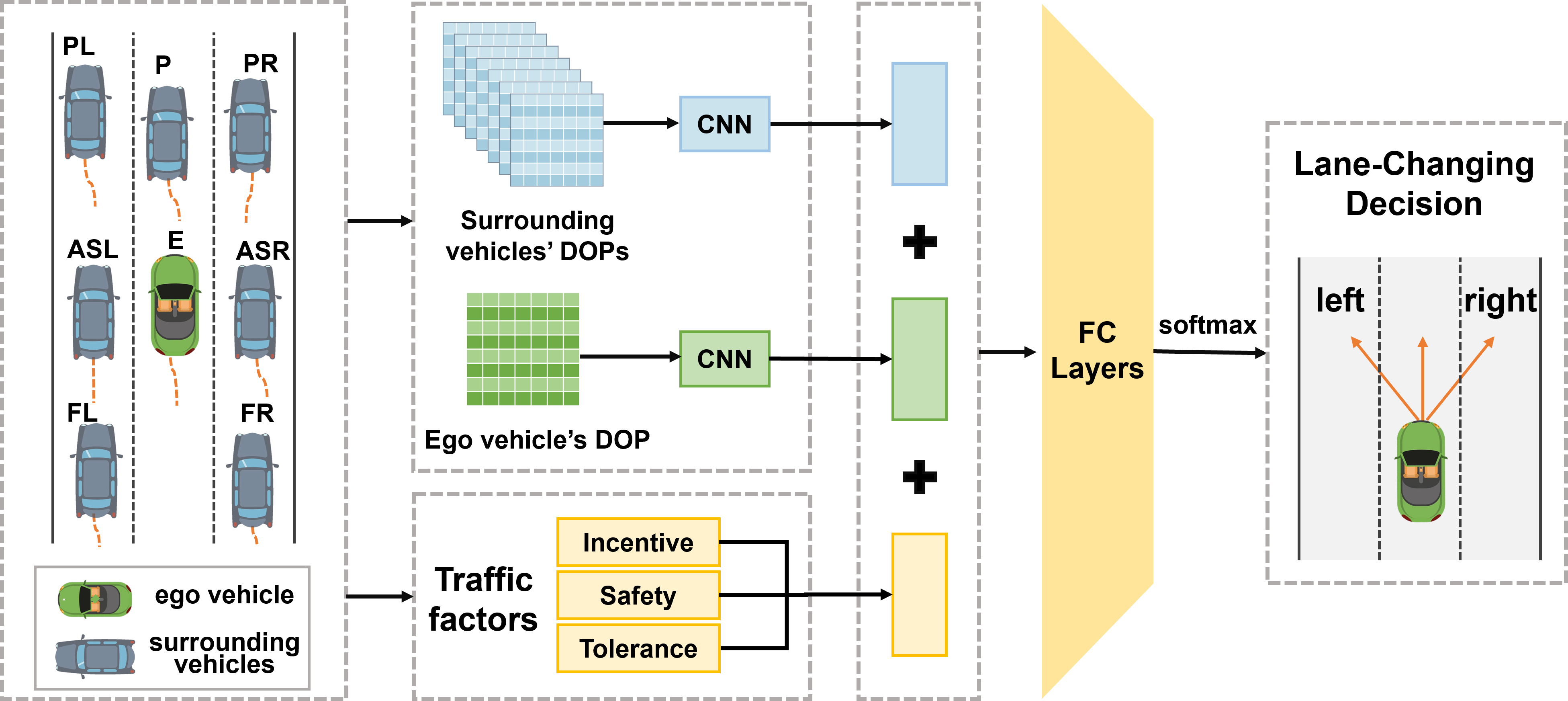}
\caption{\label{fig:DOP-LC}The architecture of LCD model considering driving style\cite{ZHANG_2023_LEARNING}.}
\end{figure}

The structural properties of deep networks facilitate the implicit mining of features. 
Taking convolutional neural network (CNN) as an example, its sensitivity to local features makes it excel in tasks that require the identification of personalized characteristics. 
\cite{ZHANG_2023_LEARNING} proposed an innovative method in which they transformed the historical trajectories of vehicles into driving operational pictures (DOP) and further converted the DOPs into continuous driving style features using CNNs. 
Based on this, they proposed a model that comprehensively considers characteristics such as incentive, safety, and tolerance, and integrated style features within it. The LCD model is illustrated in Figure \ref{fig:DOP-LC} and is represented as follows:
\begin{equation}
\label{eq_1}
    y = f_\text{LC}(F_\text{incentive}, F_\text{safety}, F_\text{tolerance}, f_\text{DOS}(\text{D}))
\end{equation}
\begin{equation}
    \label{eq_2}
    \begin{gathered}
        F_\text{incentive} = f_I(v_\text{E} - v_\text{P}, v_\text{PL} - v_\text{P}, v_\text{PR} - v_\text{P}, \\
        d_\text{PL} - d_\text{P}, d_\text{PR} - d_\text{P})
    \end{gathered}
\end{equation}
\begin{equation}
\label{eq_3}
    F_\text{safe} = f_S(d_\text{FL}, d_\text{FR}, v_\text{E} - v_\text{FL}, v_\text{E} - v_\text{FR})
\end{equation}
\begin{equation}
\label{eq_4}
    F_\text{tolerance} = f_T(d_\text{P} - v_\text{E} h_\text{E})
\end{equation}
where \(f_\text{LC}(\cdot)\) is the mapping function for LCD; 
\(f_\text{DOS}(D)\) is the implicit driving style extracted from the DOP \(D\) through CNNs; 
\(f_\text{I}(\cdot)\), \(f_\text{S}(\cdot)\), and \(f_\text{T}(\cdot)\) represent the combination functions used for the three factors, namely,  incentive, safety and tolerance, respectively; 
\(v_i\) and \(d_i\) denote the speed of vehicle \(i\) and the longitudinal distance between vehicle \(i\) and the target vehicle, respectively, where \(i \in \{\text{E, P, PL, PR, FL, FR, ASL, ASR}\}\), 
including the ego vehicle E, the preceding vehicle P in the current lane, the preceding vehicle PL in the left adjacent lane, the preceding vehicle PR in the right adjacent lane, the following vehicle  FL in the left adjacent lane, the following vehicle FR in the right adjacent lane, the alongside vehicle ASL in the left adjacent lane, and the alongside vehicle  ASR in the right adjacent lane; 
\(h_\text{E}\) is the time headway of vehicle E. 

In the decision-making process, the model incorporates not only the short-term driving style of the ego vehicle but also the driving style information of surrounding vehicles. 
Experimental results demonstrate that the model exhibits high training efficiency and predictive accuracy. 
Further analysis of misclassified samples reveals that the majority of such errors stem from the randomness of human driving behavior, indicating that the model effectively learns the general behavioral patterns of human drivers. 
If channel attention mechanisms are implemented on individual vehicles' DOP matrices, the model facilitates dynamic interaction characterization.

In various LC scenarios, researchers prefer models are capable of temporal prediction and memory retention, which is why recurrent neural networks (RNNs), particularly long short-term memory (LSTM) networks, are frequently chosen as the foundational architecture for LC models. 
Various LCD models have been developed based on improved LSTMs, such as Multi-LSTM \cite{TANG_2020_DRIVER}, attention-enhanced bidirectional multilayer residual LSTM \cite{WU_2022_DRIVER}, LSTM based on conditional random field \cite{SONG_2021_LANE}, and personalized LSTMs \cite{YE_2022_TIME}, and so on.
In direction convolutional LSTM networks\cite{ZHAO_2022_DIRECTION}, the spectrum of vehicle’s lateral trajectory proves to be a critical input feature.
When dealing with multi-dimensional features, heterogeneous data fusion models can provide more accurate predictions of LC intentions. 
For instance, \cite{YU_2021_DEEP} input driving trajectory datasets and natural driving datasets separately into LSTM networks and fused the intention probability matrices at the decision level using a mean rule to obtain a comprehensive intention. 
Under the mean rule, the output intention probability matrix of the model satisfies the following formula:
\begin{equation}
\label{eq_mean_rule_1}
    p_j^{fusion}(n) = \frac{m_j(n)}{\sum_{j=1}^{c} m_j(n)}
\end{equation}
\begin{equation}
\label{eq_mean_rule_2}
    m_j(n) = \frac{1}{k} \sum_{i=1}^{k} p_{ij}(n)
\end{equation}
where \(p_j^{fusion}(n)\) represents the fused probability of intention \(j\) for the \(n\)-th sample;
\(m_j(n)\) is the mean of the intention \(j\) for the \(n\)-th sample;
\(p_{ij}(n)\) represents the probability of intention\(j\) for  the \(n\)-th sample trained by predictive module \(i\);
\(k\) is the number of networks, with \(k=2\);
\(c\) is the number of LC intentions, with \(c=3\).

Beyond temporal dependency modeling, capturing spatial interaction dynamics in LC scenarios constitutes a critical component of DL-based LCD frameworks.
The application of graph theory enables mathematical characterization of interaction dynamics between ego vehicle and its surroundings \cite{MANZOUR_2024_VEHICLE}. Recognizing this analytical potential, \cite{CHEN_2025_SPATIOTEMPORAL} innovatively employed graph attention to model LC interactions.
\cite{GAO_2025_STMF-IE} proposed a dual-attention encoding architecture that synergistically integrates:
(1) Self-attention mechanisms to extract multi-scale temporal dependencies from historical trajectory sequences;
(2) Cross-attention modules to model interaction patterns between the ego vehicle and its neighbors, thereby enabling heterogeneous spatiotemporal feature fusion.

Enhancing the adaptability of LCD models across diverse driving environments remains a significant challenge, where online learning emerges as a viable real-time updating solution.
\cite{GU_2020_NOVEL} proposed an LC identification model based on deep autoencoders and an LCD algorithm leveraging XGBoost. 
Initially, the decision model is trained using offline historical data. 
Concurrently, the identification model continuously monitors driving behavior in real-time and collects new samples to facilitate online learning for the decision model. 
Additionally, indirect learning approaches can be employed. 
\cite{WANG_2021_BAYESIAN} introduced an adaptive driving decision threshold into the LSTM framework, developing an adaptive LC prediction model (ADLCM) based on surrounding vehicle dynamics.
Bayesian inference is utilized to estimate the threshold within short time intervals, thereby correcting decision errors induced by driving environments. 
Compared to the classical LSTM model, the proposed ADLCM demonstrates superior decision-making accuracy and effectively mitigates prediction biases caused by environmental variability. 
The improvement in ADLCM’s performance is particularly pronounced when the adaptability of the LSTM model is limited. 
This advancement not only challenges the deterministic nature of classical DL models but also significantly enhances the generalization capabilities.

However, LSTM-based models suffer from the issue of forgetting early information as the sequence length increases, along with slower inference speeds, which significantly limit their applicability in LCD modeling. 
The Transformer architecture presents three distinct advantages (including self-attention mechanisms for global spatiotemporal modeling, parallel processing architecture, and modular design flexibility), thereby establishing itself as an innovative computational framework for LCD.
Further studies demonstrate that the Transformer model exhibits significantly superior performance over traditional sequential models in LCD tasks. For instance, empirical research by \cite{GUO_2022_LANE} revealed that Transformer achieves high accuracy in LC prediction while reducing inference time by 42\% compared to LSTM. 
The Transformer architecture provides a novel methodological framework for research on  LCD of autonomous vehicles, particularly highlighting the unique advantages of its feature embedding module in multi-vehicle interaction scenarios. 
Notably, Transformer demonstrates higher accuracy in recognizing long-sequence LC intentions, whereas for short-sequence inputs, improved LSTM models remain a viable and effective alternative\cite{ZHAO_2024_LCRSS,LI_2024_DEEP}.

\subsection{LCD models based on RL}

RL, distinct from supervised learning, does not rely on extensive explicit labels but simulates human learning through trial and error with a reward mechanism. 
It enables autonomous learning in unknown environments, featuring both exploration and exploitation capabilities. 
This characteristic has garnered significant attention in the LCD of autonomous vehicles.

In RL, the reward function serves as a pivotal metric for evaluating the action of agents, whose objective is to optimize cumulative rewards through strategic learning.
For LC problems, researchers have integrated multi-dimensional objectives into the reward function design, particularly focusing on safety performance. 
For instance, \cite{XU_2020_REINFORCEMENT} constrained the agent’s action space to right LC, left LC, and LK,  and developed a multi-objective RL model that comprehensively considers safety, speed, and smoothness. 
\cite{LI_2022_DECISION} employed a probabilistic model with positional uncertainty to assess driving risk, then used RL or deep RL (DRL) to identify LC strategies that minimize expected risk. 
Innovatively, \cite{PENG_2022_INTEGRATED} proposed a dangerous action shielding mechanism, replacing safety design in reward functions by minimizing the Q-value of dangerous actions to enhance vehicle speed.
Furthermore, \cite{WU_2024_DEEP} incorporated energy consumption into the reward function, proposing an eco-friendly LCD based on double dueling deep Q network (D3QN), expanding the application dimensions of RL.
\cite{ZHANG_2023_MULTI} formulated multi-agent LCD as a partially observable Markov decision process (POMDP), which encodes surrounding vehicle intentions using a bird’s-eye view (BEV) representation and implicitly negotiates right-of-way demands. 
The reward function focuses on the speed performance of rear vehicles in the target lane to mitigate the negative LC impacts.

Whether evaluating the effectiveness of human strategies or considering the complex traffic environment of autonomous and human-driven vehicles, RL systems should incorporate continuous guidance from driver data. 
Accurately establishing reward functions and loss functions remains challenging. 
%\cite{WU_2022_SAFE} established that human-demonstration-aided RL simultaneously improves safety and reduces traffic disturbances.
\cite{WU_2022_SAFE} demonstrated that incorporating human demonstrations into RL frameworks can not only enhance safety but also mitigate traffic disturbances.
\cite{HUANG_2024_HUMAN} designed behavioral rewards based on offline human prior knowledge, while applying expert online guidance to twin delayed deep deterministic policy (TD3) gradient, proposing a human knowledge-enhanced algorithm. 
Unlike traditional manual reward function design, \cite{XU_2023_DRVING} developed a maximum entropy inverse RL (IRL) model for lane selection based on five features: safety, efficiency, comfort, collaboration, and lane priority. 
\cite{LIU_2022_INVERSE} identified driving styles through clustering analysis of driving behaviors and utilized IRL to train LCD behaviors from grouped demonstrations. 
These studies indicate that IRL provides a crucial technological approach for enhancing the anthropomorphic behavior of autonomous vehicles.

\subsection{Data-physics hybrid model}

The data-physics hybrid approaches represents a highly promising technical method, capable of ensuring high precision and interpretability, as well as offering improved adaptability and generalization capabilities. 
Physics-based methods remain effective as they rely on known system characteristics, particularly in scenarios with limited data availability. 
On the other hand, data-driven methods leverage large-scale datasets and identify complex nonlinear relationships, endowing models with dynamic adaptability. 
In hybrid approaches, the physics-based component establishes a stable behavioral foundation, while the data-driven component enables the model to adapt to diverse scenarios. 
The hybrid approach not only reduces learning costs but also enhances the transferability of model.

A simple yet effective framework incorporates high-level features derived from rules, along with vehicle-specific features, into the model. 
For instance, \cite{WANG_2022_INTELLIGENT} formulated fuzzy rules to translate driving environment information into LC feasibility. 
Subsequently, LC feasibility and corresponding vehicle trajectories were designed as input variables for an LSTM neural network. 

Notably, the mixture of experts (MoE) framework can serve as an adaptive switch between rule-based policies and data-driven networks. 
\cite{YAO_2025_MIXTRUE} proposed a hybrid architecture where intelligent driver model (IDM) and minimizing overall braking induced by lane change model (MOBIL) function as heuristic experts to complement DRL modules. 
Through a gating mechanism, these rule-based experts are dynamically activated during emergencies, upholding decision-making intelligence.

\begin{figure*}
\centering
\includegraphics[width=0.8\linewidth]{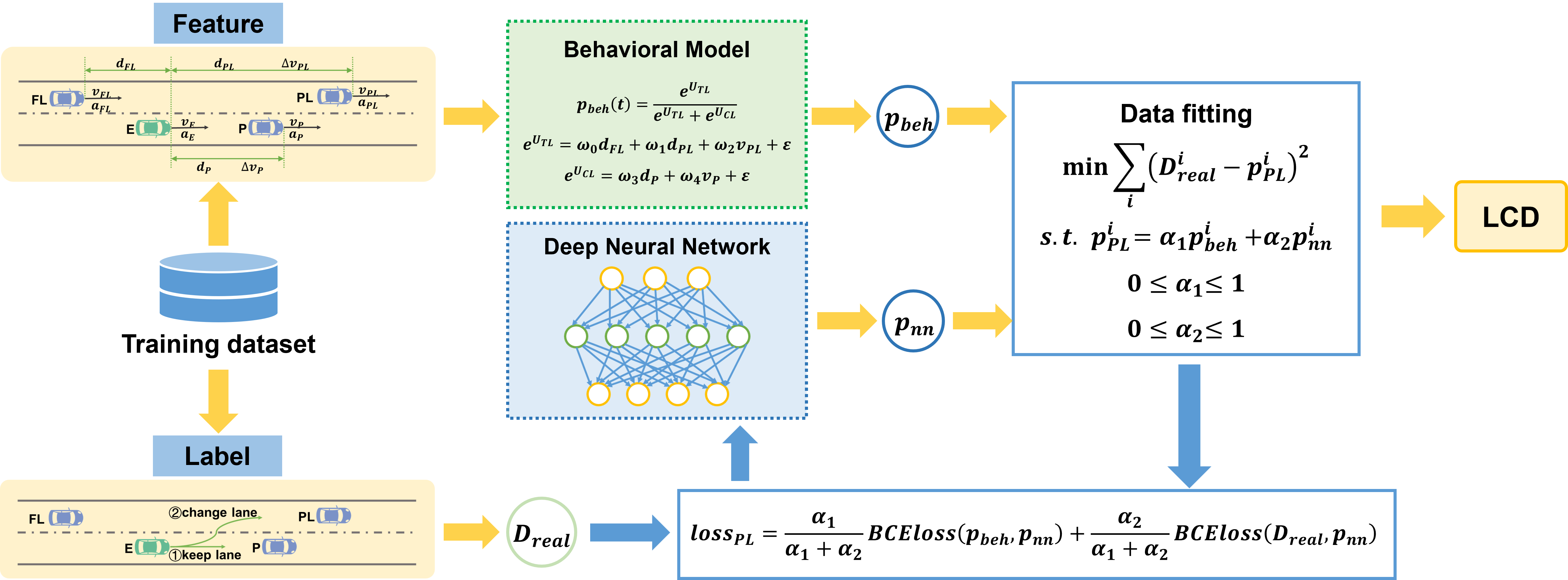}
\caption{\label{fig:parallel_model}The parallel network with dynamical weights\cite{HAN_2024_MODELING}.}
\end{figure*}

Moreover, parallel network architectures represent a common hybrid modeling framework. 
In this regard, \cite{HAN_2024_MODELING} introduced an innovative approach that embeds discrete choice models into data-driven neural network models and dynamically adjusts the weights of both components, as illustrated in Figure \ref{fig:parallel_model}. 
The optimization problem for dynamic weights is formulated as follows:
\begin{equation}
\label{eq_dynamic_weight_optimization}
    \begin{aligned}
        & \min \sum_i (D_{\text{real}}^i - p_{\text{PL}}^i)^2 \\
        & \text{s.t. } p_{\text{PL}}^i = \alpha_1 p_{\text{beh}}^i + \alpha_2 p_{\text{nn}}^i \\
        & \quad 0 \leq \alpha_1 \leq 1 \\
        & \quad 0 \leq \alpha_2 \leq 1
    \end{aligned}
\end{equation}
where \( D_{\text{real}}^i\) represents the true LCD label of the \(i\)-th sample;
\(p_{\text{PL}}^i\), \(p_{\text{beh}}^i\), and \(p_{\text{nn}}^i\) denote the predicted outputs of the parallel model, behavior model, and neural network model for the \(i\)-th sample, respectively;
\(\alpha_1\) and \(\alpha_2\) are dynamic weights.
Accordingly, the loss function for the dynamic weights, denoted as \(\text{loss}_{\text{PL}}\), can be formulated as:
\begin{equation}
\label{eq_loss_function_parallel}
    \text{loss}_{\text{PL}} = \frac{\alpha_1}{\alpha_1 + \alpha_2} \, \text{B}(p_{\text{beh}}, p_{\text{nn}}) + \frac{\alpha_2}{\alpha_1 + \alpha_2} \, \text{B}(D_{\text{real}}, p_{\text{nn}})    
\end{equation}
where \(\text{B}(\cdot)\) represents the binary cross-entropy loss. 
Experimental results demonstrate that the method of dynamically adjusting weights significantly improves parallel model performance, validating its accuracy in practical applications.

Although hybrid modeling approaches enhance the performance of LCD models, it increases complexity, requiring more expertise and computational resources for modeling and training. 
Concise expressions derived from physical equations or optimization theory provide interpretable benchmark references, enabling real-time capabilities of LCD in resource-constrained scenarios.

\section{The framework of data-driven LCD models}
\label{Principles of DDLC models}

\begin{figure*}
\centering
\includegraphics[width=0.7\linewidth]{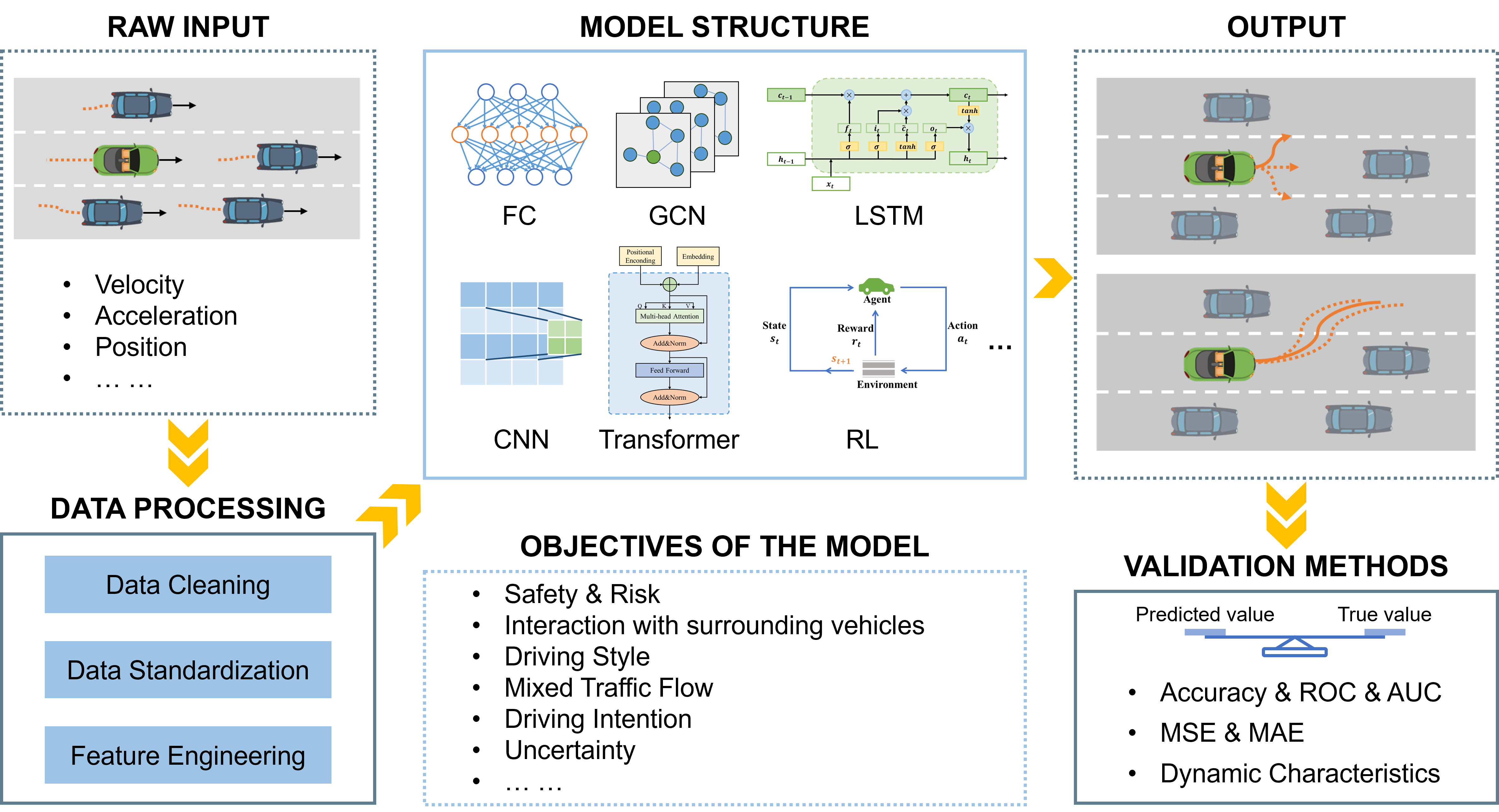}
\caption{\label{fig:framework-DDLC}Framework of data-driven LC models.}
\end{figure*}

Data-driven methods utilize field data to train and test the models, with the aim of discovering the mapping relationship between inputs and outputs.  
Figure \ref{fig:framework-DDLC} illustrates the basic framework of a data-driven LCD model, which primarily includes the data preprocessing, inputs and outputs, model structure, the objectives of the model, and validation methods. 
%The subsequent sections of this chapter will provide a systematic discussion on these key components.

\subsection{Data sources}
From the perspective of data dependency in research methodologies, the supervised learning approaches, encompassing both traditional ML and DL, exhibit high sensitivity to the scale and quality of datasets. 
Due to the challenges in acquiring naturalistic trajectory data, early vehicle trajectory datasets predominantly relied on data collected from driving simulators, supplemented by a limited amount of data obtained from actual on-road driving tests.

With advancements in technologies, the Next Generation Simulation (NGSIM) project, led by the U.S. Federal Highway Administration, emerged as an early benchmark dataset in traffic flow studies for its large-scale microscopic vehicle trajectory data \cite{FHWA_2006_NGSIM}. 
Nevertheles, the NGSIM dataset also presents several issues: 1) short collection durations; 2) limited coverage of road segments; 3) notable data noise; and 4) driving behaviors that are significantly influenced by regional factors. 

Subsequently, RWTH Aachen University (Aachen, Germany) released the HighD dataset, which includes 60 trajectory datasets collected from four observation sections near Cologne, covering an area of 420 meters\cite{krajewski_2018_HIGHD}.
Compared to NGSIM, the HighD dataset offers advantages such as a larger sample size, higher measurement precision, and smoother trajectories. However, it is worth noting that this dataset contains a substantial proportion of trucks, which may impact the generalization capabilities of models for small vehicles.

In recent years, newly released diverse datasets are pushing the boundaries of research: the Ubiquitous Traffic Eyes dataset (\url{https://seutraffic.com}) released by Southeast University (Nanjing, China); the wide-area trajectory dataset TJRD TS (\url{https://tjrdts.com}) from Tongji University (Shanghai, China)\cite{WANG_2023_WIDE_AREA}; and the INTERACTION dataset (\url{http://interaction-dataset.com}), which encompasses various traffic participants from multiple countries and scenarios.
The multimodal nature of these datasets (integrating visual, radar, and GNSS data) provides a richer feature space for developing robust LCD models. 
However, the computational costs of data cleaning remain critical challenges.

From a theoretical perspective, the training framework of RL exhibits less dependency on labeled data, as it can explore optimal strategies through self-guided interactions in simulation platforms or real-world environments. 
Currently, widely used traffic flow micro-simulation platforms include SUMO, High-env, Vissim, co-simulation platforms like CarSim/PreScan/Simulink, and hardware-in-the-loop platforms. Notably, the role of supervised learning models in simulation environments is evolving. While early studies predominantly used them as trajectory generators, subsequent research has integrated them into closed-loop control systems (e.g., SUMO TraCI interface) to achieve real-time simulation of human driving styles.

\subsection{Data preprocessing methods}

The efficacy and accuracy of data-driven models are largely contingent upon data quality. 
Common issues in vehicle trajectory datasets include insufficient data volume, noise, errors, and discrepancies in the distribution between training and testing sets, all of which can lead to significant deviations between model estimates and real values. 
Consequently, preprocessing raw data is required. 
Table \ref{tab:data process} enumerates several commonly used vehicle trajectory data processing methods as documented in the literature.

Data imbalance in vehicle trajectory datasets is manifested by the fact that LC samples are often significantly fewer than LK samples.
In scenarios of discretionary LC, the cost of misclassifying an LC sample may be substantially higher than that of misclassifying an LK sample, which can bias the model toward favoring LK strategies.
Sampling is  a direct and effective approach to address data imbalance. 
\cite{SHI_2021_IMPROVED} proposed a hierarchical oversampling bagging method based on the grey wolf optimizer (GWO) algorithm and synthetic minority over-sampling technique (SMOTE), generating more informative minority instances. 
By testing various sampling ratios, \cite{DAS_2020_DETECTING}  indicated that 1:1 sampling ratio is optimal for detecting LC maneuvers. 
Moreover, under imbalanced datasets, model outputs may exhibit a bias toward negative samples, and adjusting classification thresholds can mitigate this issue. 
Additionally, integrating diverse models and generating new data samples can further balance the dataset.

Given that trajectories encompass both temporal and spatial information, determining the exact moment when a driver decides to change lane is inherently challenging. 
Typically, changes in lateral position, speed variations, automatic labeling methods \cite{SCHEEL_2022_RECURRENT}, and wavelet transform techniques \cite{ALI_2023_calibrating} with noise reduction capabilities serve as judgment criteria. 
Calculating the start time \(t_s\) and end time \(t_e\) of LC process based on lateral displacement is the most commonly approach. For smoothed data, the following rules can be applied to determine \(t_s\) and \(t_e\):
\begin{equation}
\label{strat_time}
    t_s = \max \{ t \, | \, t < t_{LC} \text{ and } \Delta y_i(t) \Delta y_i(t - \Delta t) < 0 \}
\end{equation}
\begin{equation}
\label{end_time}
    t_e = \min \{ t \, | \, t > t_{LC} \text{ and } \Delta y_i(t) \Delta y_i(t - \Delta t) < 0 \}
\end{equation}
\begin{equation}
\label{delta_y}
    \Delta y_i(t) = y_i(t + \Delta t) - y_i(t)
\end{equation}
where, \(t_{LC}\) is the time when the lane ID changes;
\(y_i(t)\) is the lateral position of vehicle \(i\) at time \(t\);
and \(\Delta t\) is the time interval.

In addition to the constant time window method for extracting LC trajectory, we can also use the method in \cite{CHAUHAN_2022_understanding}, which requires the absolute of derivative of the cumulative lateral speed and time curve to meet the following conditions:
\begin{enumerate}
    \item \(\left| \frac{d \sum v}{dt} \right| \leq \delta\);
    \item \(\text{sign}\left(\frac{d\left| \frac{d \sum v}{dt}  \right|}{dt} \right) \text{ changed} \quad \text{or} \quad \frac{d \left| \frac{d \sum v}{dt} \right|}{dt} = 0\).
\end{enumerate}
where, \(\sum v\) is the cumulative lateral speed; \(\text{sign}(\cdot)\) represents sign function.

Numerous studies on driving behavior primarily examine successful LC, often neglecting the critical instances of failed attempts \cite{XING_2025_FAILED}. These failures, influenced by traffic flow, vehicle dynamics, road conditions, and driver judgment, provide valuable insights into the complexities of LC behavior. Analyzing such cases may enhance risk assessment, decision-making frameworks, and a deeper understanding of driving behavior.

\begin{table*}[!ht]
\caption{\label{tab:data process}Trajectory data processing methods.}
\centering
\begin{threeparttable}
\begin{tabular}{p{1.5cm}p{1.5cm}p{10cm}p{3cm}}
\hline
\toprule 
\textbf{Literature} & \textbf{Data source} & \textbf{Data preprocessing and filtering} & \textbf{Ratio of LC data} \\ \midrule
Xie et al.\cite{XIE_2019_DATADRIVEN} & NGSIM & 1) Olny cars are selected;
2) Vehicles present on lanes 1 to 4;
% 3) Excluding vehicles that perform multiple lane changes or traverse multiple lanes;
3) Vehicles performing multiple lane changes or crossing multiple lanes are excluded;
4) Duration more than 10 seconds. & 1078 LC \& 1120 LK \\
Zhang et al.\cite{ZHANG_2023_LEARNING} & HighD & 1) The automatic labeling method is implemented to identify LCD moments;
2) Vehicles staying in the same lane for more than 12\(s\) seconds are classified as LK vehicles;
% 3) a 16-fold replication of the LC data. 
3) LC event augmentation employs 16-fold replication. 
& 4099 LCL \& 5150 LCR \\
Tang et al.\cite{TANG_2020_DRIVER} & NGSIM & 
1)Smoothing trajectories by symmetric exponentially weighted moving average filter;
% 2) LC start point where the change in lateral velocity exceeds a predefined threshold;
% 3) Manual verification.
2) LC start point is defined as the place where the lateral velocity change exceeds a predefined threshold;
3)Manual verification confirms the results.
& 1363 LCL \& 280 LCR \& 1687 LK \\
% Han et al. & Data in Nanjing, China & Undersampling. & 1179 LC \& 1122 LK \\
% Gao et al.\cite{GAO_2023_DUAL} & NGSIM \& HighD & 1) Unification of position units;
% 2) Introduction of virtual vehicles when the number of surrounding vehicles is fewer than five. & - \\
\bottomrule
\end{tabular}
\begin{tablenotes}
       \footnotesize
       \item[1] Note: LCR(LC to the Right Lane), LCL(LC to the Left Lane).
  \label{tab:example}
\end{tablenotes}
\end{threeparttable}
\end{table*}

\subsection{Inputs and outputs}

% The inputs of LCD models directly influence the model’s information processing approach.
Input data structures predispose processing layer selection without strictly dictating it.
% For instance, in applications considering temporal dependencies, the inputs can be structured as sequences.
For instance, sequential inputs tend to favor attention mechanisms in temporal processing, although they are not strictly necessary.
Meanwhile, the complexity of LCD models primarily stems from conflicting objectives and numerous influencing factors. 
Existing studies have selected a diverse array of input variable types, as illustrated in Table \ref{tab:content(in_out)}.
In particular, \cite{HUANG_2018_CAPTURING} proposed an LC model based on social force theory, with inputs defined as follows:
\begin{equation}
    \label{eq_input_1}
    \begin{gathered}
        I = \{\mathbf{TD}_i(t); L_i; \mathbf{s}_i(t-n\tau); L_j; \Delta \mathbf{s}_{(j,i)}(t-n\tau)\}, \\
        n = 0, 1, \ldots, N, \quad j \in \{L, F, TL, TF\}\\
    \end{gathered}
\end{equation}
\begin{equation}
    \label{eq_input_2}
    \mathbf{TD}_i(t) = \{\Delta \mathbf{D}_i(t), \Delta T_i\}
\end{equation}
\begin{equation}
    \label{eq_input_3}
    \Delta \mathbf{D}_i(t) = \mathbf{D}_i(t) - \mathbf{s}_i(t)
\end{equation}
\begin{equation}
    \label{eq_input_4}
    \Delta T_i = T_i - t
\end{equation}
\begin{equation}
    \label{eq_input_5}
    \mathbf{s}_i(t) = \{x_i(t), y_i(t)\}
\end{equation}
\begin{equation}
    \label{eq_input_6}
    \Delta \mathbf{s}_{(j,i)}(t) = \begin{cases} \mathbf{s}_j(t) - \mathbf{s}_i(t) & \text{if } j \text{ exists} \\ \infty & \text{else} \end{cases}
\end{equation}
\begin{equation}
    \label{eq_input_7}
    \mathbf{s}_j(t) - \mathbf{s}_i(t) = \{x_j(t) - x_i(t), y_j(t) - y_i(t)\}
\end{equation}
where,
\(L_i\) and \(L_j\) represent the lengths of vehicles \(i\) and \(j\), respectively;
\(\Delta \mathbf{D}_i(t)\) denotes the relative distance of vehicle \(i\) to the destination;
\(\Delta T_i\) indicates the relative expected time for vehicle \(i\) to reach the destination;
\(\mathbf{D}_i(t)\) represents the relative position of vehicle \(i\) at time \(t\)  to the designated destination;
\(T_i\) is the expected time for vehicle \(i\) to reach the designated destination;
\(x_i(t)\) and \(y_i(t)\) denote the vertical and horizontal positions of vehicle \(i\)  at time \(t\), respectively;
\(\tau\) is the time interval for each step;
%for calculating speed, 
and \(L, F, TL, TF\) denote the lead vehicle, following vehicle in the current lane, lead vehicle and following vehicle in the target lane, respectively.

The inputs listed in Table \ref{tab:content(in_out)} are broadly categorized into vehicle state, driver, and environment characteristics. 
Vehicle state attributes (such as speed, acceleration, and steering angle) reflect real-time dynamics. 
Driver habits, often highly personalized, can be inferred from historical trajectories. 
The traffic environment, defined by inter-vehicle relationships (e.g., gaps, relative speeds, and surrounding vehicles), aids in understanding driver intent. 
The role of the traffic environment in supporting human driving has been well-documented.
Additionally, road conditions, weather, traffic flow, and vehicle types significantly influence LCDs of drivers. 

Excessive features introduce redundancy and reduce computational efficiency, making the selection of concise, effective inputs a critical focus in data-driven LCD modeling.

\begin{table*}[!ht]
\caption{\label{tab:content(in_out)}Literature, inputs, outputs,objectives and validation methods of models.}
\centering
\begin{threeparttable}
\begin{tabular}{p{1.1cm}p{1.7cm}p{1cm}p{3cm}p{2.5cm}p{2cm}p{4cm}
}
\hline
\toprule
\textbf{Category} & \textbf{Method} & \textbf{Literature} & \textbf{Inputs (factors or variables)} & \textbf{Outputs}  & \textbf{Objectives} & \textbf{Validation methods} \\ 
\midrule
Supervised learning & Ensemble learning & Hou et al. \cite{HOU_2015_SITUATION} & Speed, relative speed, spacing & \{LC, KL\} & Safety & Acc, TPR, TNR \\ 
 & HMM-BF & Li et al. \cite{LI_2016_LANE} & Steering angle, lateral acceleration and yaw rate & \{LC, KL\} & Intention recognition & Average values of the output distribution, ROC, Recognition time  \\
 & HMM & David et al. \cite{DAVID_2022_STUDY} & State of ego/surrounding vehicles and driver’s operational information & \{LCL, LCR, LK\} & Driving behavior recognition & Acc, Recall, FPR \\ 
 % & Dynamic Bayesian Network & Jiang et al. \cite{JIANG_2023_VEHICLE} & Lateral speed, direction angle & Position & Driver uncertainty, interaction & Log probabilities of intention, maximum error, mean error, standard deviation\\ 
 & DBN+LSTM & Xie et al. \cite{XIE_2019_DATADRIVEN} & Speed, relative speed & \{LC, KL\}; position & LCD prediction Acc  & Acc, MSE; MSE \\ 
 & CNN & Zhang et al. \cite{ZHANG_2023_LEARNING} & Spacing, relative speed & \{LCL, LCR, LK\} & Driving style & Acc, TPR, TNR, clustering DOPs, speed change rate, TTC \\ 
 % & LSTMs & Zhang et al. \cite{ZHANG_2019_simultaneous} & time-sequence historical position information of subject vehicle and six surrounding vehicles & Two-dimensional positions & Model integrated LC and LK & MSE, lateral accuracy, longitudinal error, mixed gap error, reasonable prediction accuracy, transferability analysis  \\
 & Multi-LSTM & Tang et al. \cite{TANG_2020_DRIVER} & Time series of speed & \{LCL, LCR, LK\} & Intention & Acc \\ 
 & Transformer & Guo et al. \cite{GUO_2022_LANE} & Sequence of speed, yaw rate, acceleration, relative speed, and distance & \{LCL, LCR, LK\} & Driving scenarios & Time, PR, AUC, F1 \\
 % & LSTM-GP-IMM & Chen et al. \cite{CHEN_2024_LANE} & Historical trajectory & Trajectory & Driving style, uncertainty, interaction & Confidence interval, RSME, MAE, final displacement error\\ 
 & Parallel learning-based model & Han et al. \cite{HAN_2024_MODELING} & Speed, acceleration, distance, relative speed; sequence of position, speed, output of LCD, 
 & \{LC, KL\}; position when LC maneuver ends & Interpretability, transferability & Acc, recall, precision, F1, ROC, AUC; MAE\\
\midrule
RL & Multiobjective approximate policy iteration & Xu et al. \cite{XU_2020_REINFORCEMENT} & Lane index, speed, distance & \{LCL, LCR, LK\} & Safety, speediness, smoothness & Average velocity, minimum vehicle distance, reward, accumulative reward, real-Time Experiments \\
 & DRL with discrete actions & Li et al. \cite{LI_2022_DECISION} & Relative distance (both longitudinal and lateral), yaw angle, and yaw rate & Discrete steering action & Driving risk, uncertainty & Score(e,g. traveled distance), relative change rate, risk assessment, motion path \\
 & Maximum entropy IRL & Xu et al. \cite{XU_2023_DRVING} & Gap, speed, relative speed, lateral acceleration, jerk, and lane priority &  \{LCL, LCR, LK\} & Safety, efficiency, cooperativity, human-like & RMSE, risk \\
 & D3QN+DDPG & Peng et al. \cite{PENG_2022_INTEGRATED} & Speed, acceleration, lane index and the information of the following vehicles; speed, acceleration and headway & \{LCL, LCR, LK\}; acceleration & Scene complexity, coupling, safety, efficiency, comfort & Spatio-temporal trajectory, efficiency improvement in different scenes, LC frequence, TTC, average acceleration   \\

\bottomrule
\end{tabular}
\begin{tablenotes}
       \footnotesize
       \item[1] Note: Acc(Accuracy), TPR(True Positive Rate), TNR(True Negative Rate), FPR(False Positive Rate), ROC(Receiver Operation Characteristic curve), PR(Precision-Recall curve), AUC(Area Under Curve), MSE(Mean Squared Error), RMSE(Root Mean Square Error), MAE(Mean Absolute Error).
  \label{tab:example}
\end{tablenotes}
\end{threeparttable}
\end{table*}

\subsection{Model structure}

% In the domain of ITS, DL approaches integrate multi-source data and then learn complex LC behavioral patterns, thereby approximating human driving decision processes \cite{YU_2021_DEEP,ZHANG_2023_LEARNING,WANG_2021_BAYESIAN}. 
In ITS, DL methodologies effectively fuse heterogeneous data sources to model intricate LC behavioral patterns, closely emulating human driver decision-making processes \cite{YU_2021_DEEP,ZHANG_2023_LEARNING,WANG_2021_BAYESIAN}.
Nevertheless, prevailing DL methods encounter several technical limitations: (1) insufficient causal relationship modeling capacity, impeding capture of causal mechanisms; (2) constrained model generalization capability, resulting in poor performance in unencountered driving scenarios \cite{GU_2020_NOVEL}; and (3) lack of model interpretability, which restricts their application in safety-critical scenarios. 
Despite these limitations, these methods retain significant value in accurate LCD estimation and behavior prediction, particularly in the current stage of incomplete vehicle-to-everything (V2X) technology maturation.

RL, able to achieve objective-driven, offers a highly adaptive solution for autonomous LCD through unsupervised exploration and dynamic reward mechanisms\cite{XU_2020_REINFORCEMENT}. 
With the evolving complexity of real traffic systems (including fluctuations in human-machine interaction ratios and increasing heterogeneity in driving styles), the research focus has shifted towards achieving Pareto optimality among safety, traffic efficiency, and social coordination while maintaining human-like behavioral modeling fidelity. 
The primary challenges in RL practical applications lie in: (1) achieving smooth policy transfer from simulation to real-world scenarios, and (2) effectively addressing environmental distribution discrepancies and dynamic disturbances. 
These challenges remain core constraints impeding RL technology implementation.

While single-algorithm models possess structural simplicity and implementation ease, their generalization performance is limited in complex dynamic traffic scenarios. 
To overcome this issue, recent studies are progressively transitioning toward multi-module fusion and ensemble learning architectures, enhancing accuracy and generalization through complementary advantages \cite{HAN_2024_MODELING}. This approach also presents novel solutions for agent-driver coordination in real traffic environments. 
However, several challenges persist: (1) system redundancy and complexity, where feature overlap between modules increases data processing burdens, and parallel model execution leads to computational resource waste and conflicting LCD suggestions;
(2) heterogeneous integration, where feature space and decision mechanism disparities among models affect fusion effectiveness, necessitating adaptive weight allocation strategies;  
and (3) solution space explosion, where partial observability, long-term decision dependencies, multi-agent coordination, environmental asymmetry, and high stochasticity collectively constitute a vast solution space, imposing more stringent requirements on algorithm design and optimization.

\subsection{Objectives of LCD}

Data-driven mechanisms are driving models to progressively explore behavioral patterns in driving data, thereby enabling the effective utilization of multi-source information. 
The core objective of LCD models is to achieve safe, efficient, and human-like driving decisions through data-driven approaches. 
Illustrated in Table \ref{tab:content(in_out)}, supervised learning leverages labeled data to learn human driving behaviors, while RL guides decision-making through safety-related reward functions \cite{XU_2020_REINFORCEMENT,WU_2024_DEEP}, supplemented by risk-based rule constraints that enhance model reliability \cite{ZHANG_2023_MULTI,PENG_2022_INTEGRATED}. 
Collectively, these approaches underscore the paramount importance of safety in LCD. 
Concurrently, research has conducted multi-scenario analysis of LCD behavior from the perspectives of traffic flow efficiency, personalized driving styles, and interactive behaviors.

As research into micro LCD mechanisms, human driving behaviors, and driving scenarios for AVs deepens, the focal point has gradually shifted from ``safety in LC" to ``human-like LCD behavior" \cite{HUANG_2024_HUMAN,XU_2023_DRVING,ZHAO_2023_INTERACTION}.
This transition reflects a consensus within the academic community regarding the environment of AV: in the foreseeable future, AVs will coexist with human-driven vehicles in complex environments for the long term. 
The underdevelopment of vehicular communication technology and the immaturity of human-machine collaboration mechanisms render the optimization of information-sharing efficiency and the integration of human driving flexibility with AV computational advantages critical research directions.

As the penetration rate of AVs gradually increases, traffic behavior patterns will evolve from ``human-behavior dominant" to ``efficient collaborative optimization''.
In the low penetration case, models emphasize learning human driving behaviors and reasoning methods.
Conversely, in the high penetration case, models focus more on optimizing overall traffic flow efficiency. 
This evolutionary not only facilitates the seamless integration of AVs into existing traffic systems but also lays a vital foundation for achieving efficient, coordinated, and safe road traffic environments in the future.

In human-vehicle-environment multi-factor systems, the pursuit of multiple optimization objectives while ensuring human-like LCD is a prominent research direction. 
However, this may lead to high computational costs and local optima in data-driven methods. 
Therefore, it is essential to balance model efficacy with computational cost in practical applications.

\subsection{Validation methods}

% Model validation, which assesses the effectiveness of data-driven LC models through accuracy, safety, and reliability metrics, constitutes a critical component in the development process. 
% The existing validation framework primarily encompasses two dimensions: prediction accuracy and vehicular characteristic performance. 
% In terms of prediction accuracy, LCD models predominantly employ performance metrics such as Accuracy, True Positive Rate (TPR), and Receiver Operating Characteristic (ROC) Curve to evaluate decision-making reliability. 
% Conversely, LCI models typically utilize error-based metrics including Mean Squared Error (MSE), Mean Absolute Error (MAE), and Root Mean Square Percentage Error (RMSPE), focusing on predictive capabilities regarding key parameters such as position, velocity, acceleration, and inter-vehicle spacing. 
% The vehicular characteristic performance dimension emphasizes driving style, efficiency, safety, and comfort, evaluating the model’s ability to achieve desired behavioral attributes.

% Nevertheless, the current validation process presents several limitations: (1) the absence of benchmark models hinders objective cross-model comparisons and performance evaluations; and (2) the lack of standardized evaluation protocols complicates systematic assessment of LC performance and its influencing factors.

% Future research should prioritize the establishment of more systematic and comprehensive validation standards, developing multi-dimensional evaluation index systems.

In terms of prediction accuracy, detailed in Table \ref{tab:content(in_out)}, the LCD model typically employs performance metrics such as accuracy, true positive rate, and receiver operating characteristic (ROC) curves to evaluate decision outcomes. 
% Conversely, the LCI often utilizes error calculation indicators like mean squared error (MSE), mean absolute error (MAE), and root mean square percentage error (RMSPE) to assess the model's predictive capabilities regarding key parameters such as position, speed, acceleration, and headway. 
%The latter category focuses on vehicle features (e.g., driving style, efficiency, safety, comfort), examining how well the model achieves these characteristics.
Regarding feature performance, the focus is typically on vehicle attributes, such as driving style, efficiency, safety, and comfort. 
% This examination evaluates the extent to which the model successfully embodies these characteristics.

However, from the perspective of the data testing, the validation process is not entirely adequate, and there still exist some issues.

\begin{itemize}
\item Lack of a unified baseline: 
%The absence of a standardized baseline for comparative models leads to inconsistencies in energy efficiency comparisons, making effective horizontal comparisons difficult.
Due to the absence of a unified benchmark model, the horizontal comparison among various models lacks a reference point, making it difficult to verify the superiority of model performance.
This lack of a unified evaluation framework can result in misunderstandings or overestimated assessments of model performance.

\item Absence of a systematic evaluation framework: 
%Current research lacks a unified standard for evaluating the performance of lane change vehicles and the factors influencing their behavior. 
LC actions are affected by various elements, including dynamic traffic flow, driver behavior, and road conditions. 
The lack of a comprehensive and systematic evaluation framework complicates the quantification of these combined and intricate effects.

\item Oversimplified performance interpretation: 
%Accuracy rate is a critical measure of a predictive model's performance—and particularly vital in imitation learning for mitigating safety hazards—it does not encompass other important factors.
The effectiveness of many data-driven LCD models is interpreted solely from the perspective of accuracy, which is rather limited and lacks in-depth exploration of the underlying mechanisms. 
%Prediction accuracy is the most direct and effective performance metric for prediction models; 
%in the context of imitation learning, it is also the most critical component in mitigating safety hazards within the system. 
While prediction accuracy is an important indicator, factors such as safety, stability, and robustness are equally crucial in real-world traffic scenarios. 
Relying solely on prediction accuracy to evaluate the effectiveness of a model fails to fully reflect its actual performance in complex traffic environments. 
%Safety, comfort, and efficiency are common evaluation metrics for LC models. However, during the LC process, safety and efficiency often act as competing factors, leading to conflicting assessment outcomes. 
%Therefore, it is essential to seek appropriate methods to reconstruct the benefit evaluation model for LC behavior.

\end{itemize}

Consequently, future research should develop a more systematic and comprehensive validation standard, incorporating multidimensional evaluation indicators to better reflect the actual effectiveness and reliability of the model in real-world applications.

\section{Opportunities and challenges}
\label{Opportunities and challenges}

As the development of artificial intelligence, researchers established various data-driven models specifically targeting LC behaviors. 
The current research focus has shifted toward modeling LC in complex scenarios while enhancing model interpretability and deepening the foresight of decision-making outcomes. 
Building upon preceding discussions, this section endeavors to explore the existing limitations and potential opportunities for LCD modeling.

\subsection{Robustness challenges}

Data-driven LCD models demonstrate a dual critical characteristic that warrants rigorous examination: the coexistence of ``high-ceiling" performance and ``low-floor" risks. 
The ``high-ceiling" denotes their capability to achieve efficient LCD behaviors under ideal conditions, while the ``low-floor" reflects suboptimal prediction accuracy and inappropriate LCD timing decisions during data distribution shifts. 
This performance volatility exacerbates concerns regarding the black-box nature of data-driven approaches, highlighting the urgent need for enhanced risk resilience and system robustness.

To address such robustness challenges, future research can focus on developing fault-tolerant mechanisms that synergistically integrate multiple learning methods. 
These mechanisms should maintain operational integrity during partial component failures, thereby enhancing model reliability. 
Notably, systematic integration of transfer learning and adversarial domain adaptation techniques shows potential to enhance model robustness against distributional shifts. 
Furthermore, implementing adaptive optimization methodologies that dynamically adjust model parameters in response to varying environmental conditions could significantly improve reliability against domain shifts and data distribution discrepancies.

\subsection{Triple Challenges of Uncertainty}

The uncertainties in LCD models can be categorized into three primary aspects:

\subsubsection{Perception uncertainty} 
It primarily stems from sensor noise and environmental variability. 
In addition, data acquisition devices and communication infrastructure under adverse weather conditions significantly affect environmental perception accuracy,  thereby increasing data noise.

\subsubsection{Behavioral uncertainty} 
Driver heterogeneity introduces multimodal decision-making patterns, where human drivers exhibit diverse behaviors even in similar scenarios. 
This variability is further complicated in real traffic environments, where surrounding vehicles' abrupt behavioral changes in response to environmental shifts create additional prediction challenges. 
Deterministic behavioral planning frameworks struggle to accommodate such human behavioral stochasticity.

\subsubsection{Model uncertainty} 
It comes from both data characteristics \cite{LI_2024_DEEPLEARNING} and model architecture. Hyperparameter selection and randomness in the training process can contribute to unstable outcomes.
Additionally, various data imperfections (such as incompleteness, noise, inconsistencies, multimodality, and long-tail distributions) can significantly impact the sensitivity of the LCD model in diverse traffic scenarios.

Current mitigation strategies employ probabilistic models and multi-module architectures. 
% However, developing a comprehensive framework with unified quantitative evaluation metrics remains crucial for effectively addressing these uncertainties and enhancing model validity and generalizability.
Emerging methodological paradigms reveal a critical shift in research focus: rather than pursuing deterministic prediction of stochastic human behaviors, quantifying the reliability of LCD models' probabilistic predictions offers greater theoretical and practical significance, particularly in enhancing model interpretability and operational transparency.

\subsection{Technical Framework Evolution}
\begin{figure*}
\centering
\includegraphics[width=0.9\linewidth]{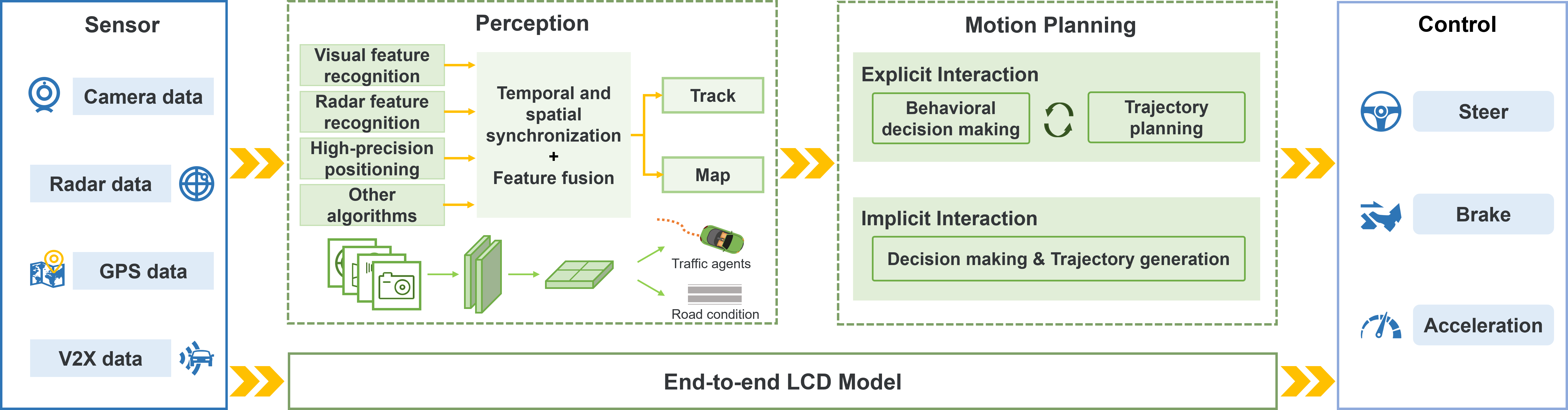}
\caption{\label{fig:end2end}Modular and end-to-end method architecture.}
\end{figure*}

Vehicle operation constitutes a complex behavior typically decomposed into perception, decision-making, and control modules in Figure \ref{fig:end2end}. 
However, hierarchical architectures suffer from a critical flaw: Decision-layer errors may propagate to planning and control components. 
The rapid advancement of large language models (LLMs) has catalyzed interest in end-to-end autonomous driving control methods, which generate control strategies directly from visual inputs. These integrated models combine multiple data-driven modules to reduce development and maintenance costs, albeit at the expense of becoming more complex ”black boxes”.

The powerful capabilities of LLMs rely on abundant datasets and powerful computational capabilities.
This raises a noteworthy question: Under conditions of limited memory capacity in these LCD models, whether forgetting phenomena will occur, particularly for marginal events such as extreme weather and sporadic incidents, remains a question. 
Additionally, whether the LCD models without rule constraints can achieve effective interaction with real traffic participants is also a topic that urgently needs to be explored.

Currently, the integration of RL with LLMs is widely regarded as an important research direction, signifying a  trend in the research of general autonomous systems. 
LLMs, endowed with extensive pre-training knowledge and superior language comprehension capabilities, can serve as effective scene interpreters, addressing inherent data efficiency challenges in RL, which primarily stem from the complexity of learning high-dimensional state representations \cite{WU_2024_RECENT}. 
However, while end-to-end models enhance efficiency by integrating intermediate processes, this also imposes stricter safety requirements on these models.

The lack of interpretability in data-driven LCD models implies potential risks.
Currently, end-to-end models for AVs are striving to incorporate natural language descriptions into the decision-making and action processes, enabling users to interact and provide feedback during model predictions, thereby continuously adjusting and optimizing the model's interpretability. 
Furthermore, data-driven LCD control may also involve ethical and legal issues, such as determining liability in the event of an accident. 
The absence of clear decision-making bases may trigger legal disputes, affecting the promotion and application of the technology.

\section{Conclusion}
\label{Conclusion}

Data-driven LCD models can well capture complex driving behaviors, as well are essential for AVs and ITSs. 
Compared to traditional rule-based approaches, data-driven models are more adaptable to diverse driving styles and environmental conditions, fully accounting for the dynamic interactions between vehicles and surrounding traffic participants.
The evolution from classical ML to advanced DL and DRL architectures has substantially increased the sophistication of these models' core mechanisms. 
However, it is necessary to conduct research on the logic of the models, in order to address the challenges related to model interpretability, generalizability, and robustness in complex traffic environments.

A fundamental insight emerges: model complexity should align with specific application objectives. 
For traffic simulation purposes, human-like imitation learning frameworks may suffice for reproducing fundamental traffic characteristics. 
Conversely, AV control demands more sophisticated models to mitigate potential risks effectively. 
To this end, the intended purpose of LCD modeling is critical for selecting appropriate methodologies.

The evolution of AV technologies highlights the necessity for optimal decision-making strategies. 
While data-driven methods present promising solutions, they encounter challenges related to safety, data efficiency, and generalizability. 
In this context, ensuring the safe and efficient integration of these strategies into AVs is paramount. 
Developing hybrid models that leverage complementary advantages may be one of the approaches to enhance the effectiveness of data-driven LCD models.
In the future, the integration of technologies such as multimodal data, LLM and RL is expected to further enhance the performance of data-driven LCD models. 
This advancement will support safe and efficient vehicle control, ensuring timely and accurate decision-making across a variety of real-world LC scenarios.

\bibliographystyle{IEEEtran}
\bibliography{DDLC_V2}

\begin{IEEEbiography}[{\includegraphics[width=1in,height=1.25in,clip,keepaspectratio]{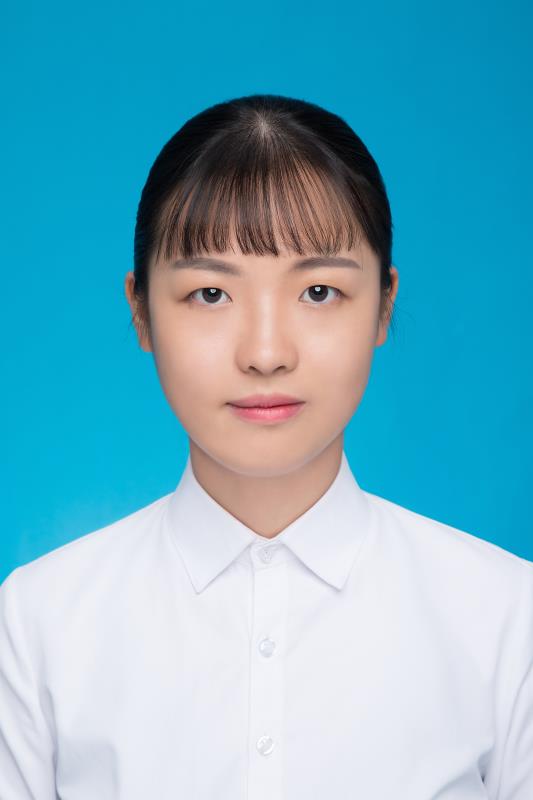}}]{Linxuan Huang} 
  received the B.S. degree in 2023. She is currently pursuing the master’s degree in Traffic and Transportation Planning and Management from Beijing Jiaotong University. Her main research interests include traffic simulation, connected vehicle technology, traffic data analysis, and data-driven traffic system modeling.
\end{IEEEbiography}

\begin{IEEEbiography}[{\includegraphics[width=1in,height=1.25in,clip,keepaspectratio]{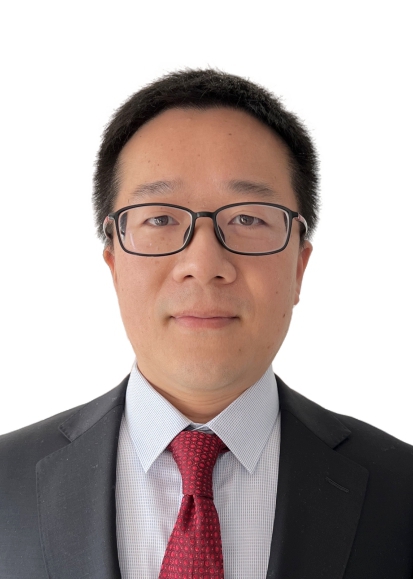}}]{Dongfan Xie} 
  received the B.S. degree in traffic engineering and the Ph.D. degree in Traffic and Transportation Planning and Management from Beijing Jiaotong University, Beijing, China, in 2005 and 2011, respectively. From 2011/5 to 2015/5, he worked as a research fellow in the School of Civil and Environmental Engineering in Nanyang Technological University, Singapore. From 2012/5 to 2016/12, he worked as a lecturer in the School of Traffic and Transportation in Beijing Jiaotong University. He is currently a Professor with the School of Systems Science, Beijing Jiaotong University. His current research interests include the traffic flow modeling, intelligent transportation system, data-driven traffic system modeling, and network traffic dynamics.
\end{IEEEbiography}

\begin{IEEEbiography}[{\includegraphics[width=1.1in,height=1.3in,clip,keepaspectratio]{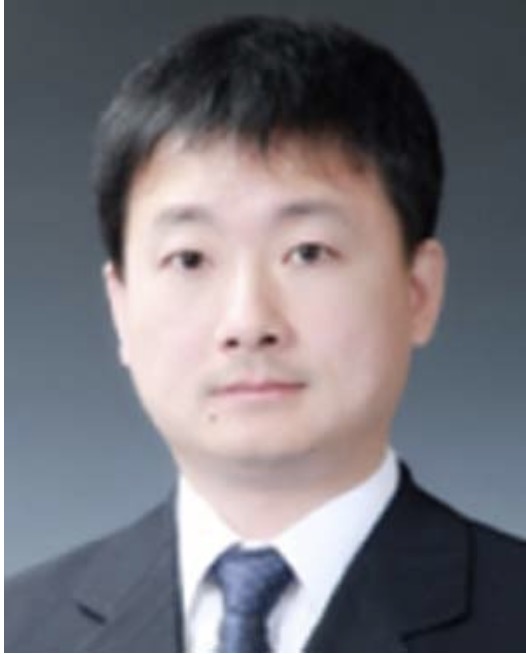}}]{Li Li} (Fellow, IEEE)) is currently a Professor with the Department of Automation, Tsinghua University, Beijing, China. He is working in the fields of artificial intelligence, intelligent control and sensing, intelligent transportation systems, and intelligent vehicles. He has authored or coauthored more than 170 SCI-indexed international journal articles and more than 70 international conference papers as a first/corresponding author. He is a Member of the Editorial Advisory Board for Transportation Research Part C: Emerging Technologies, and a Member of the Editorial Board for Transport Reviews and Acta Automatica Sinica. He is also an Associate Editor for IEEE TRANSACTIONS ON INTELLIGENT TRANSPORTATION SYSTEMS and IEEE TRANSACTIONS ON INTELLIGENT VEHICLES.
\end{IEEEbiography}

\begin{IEEEbiography}[{\includegraphics[width=1in,height=1.25in,clip,keepaspectratio]{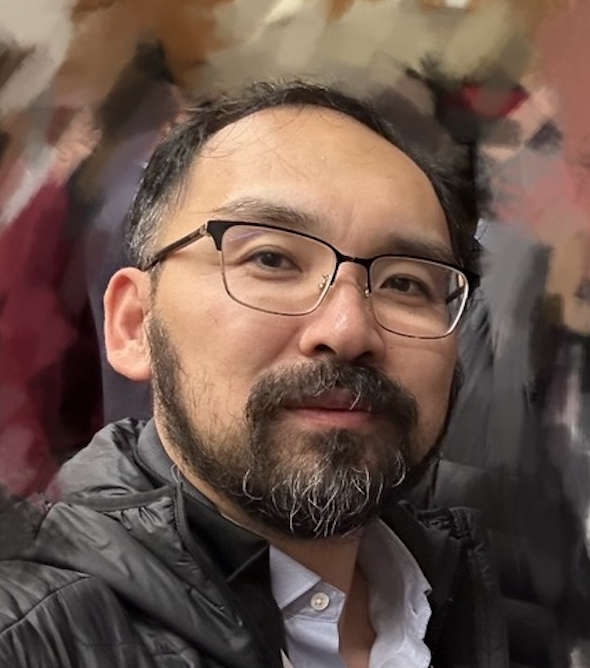}}] 
{Zhengbing He} (M'17-SM'20) received the Bachelor of Arts degree in English language and literature from Dalian University of Foreign Languages, China, in 2006, and the Ph.D. degree in systems engineering from Tianjin University, China, in 2011. He was a Post-Doctoral Researcher and an Assistant Professor with Beijing Jiaotong University, China. From 2018 to 2022, he was a Full Professor with Beijing University of Technology, China. Presently, he is a Research Scientist with the Laboratory for Information and Decision Systems (LIDS), Massachusetts Institute of Technology, USA. 

His research stands at the intersection of transportation, systems engineering, and artificial intelligence, with a focus on topics such as data-driven modeling and intelligent vehicle-enabled congestion solutions, learning-based sensing and prediction of traffic congestion and travel demand, and sustainability-oriented optimization in transportation. He has published more than 160 academic papers, with total citations exceeding 6,000. He was listed as World’s Top 2\% Scientists. He is the Editor-in-Chief of the Journal of Transportation Engineering and Information (Chinese). Meanwhile, he serves as a Senior Editor for IEEE TRANSACTIONS ON INTELLIGENT TRANSPORTATION SYSTEMS, Deputy Editor-in-Chief of IET Intelligent Transport Systems, a Handling Editor for Transportation Research Record, and an Editorial Advisory Board Member for Transportation Research Part C. His webpage is https://www.GoTrafficGo.com.
\end{IEEEbiography}

\vfill

\end{document}